# Editorial introduction: The power of words and networks

Fronzetti Colladon, A., Gloor, P., & Iezzi, D. F.

# Editorial Introduction: The Power of Words and Networks

Fronzetti Colladon, A., Gloor, P., & Iezzi, D. F

**Abstract**

According to Freud "words were originally magic and to this day words have retained much of their ancient magical power". By words, behaviors are transformed and problems are solved. The way we use words reveals our intentions, goals and values. Novel tools for text analysis help understand the magical power of words. This power is multiplied, if it is combined with the study of social networks, i.e. with the analysis of relationships among social units. This special issue of the International Journal of Information Management, entitled "Combining Social Network Analysis and Text Mining: from Theory to Practice", includes heterogeneous and innovative research at the nexus of text mining and social network analysis. It aims to enrich work at the intersection of these fields, which still lags behind in theoretical, empirical, and methodological foundations. The nine articles accepted for inclusion in this special issue all present methods and tools that have business applications. They are summarized in this editorial introduction.

"Words were originally magic and to this day words have retained much of their ancient magical power. By words one person can make another blissfully happy or drive him to despair" (Freud, 1989). Word were precious in the past and used to cast spells or talk to gods. Human change can be promoted by the use of words. By words, behaviors are transformed and problems are solved (Watzlawick, Weakland, & Fisch, 2011). Nowadays, novel tools for text analysis help understand the magical power of words. The way people express themselves can reveal their intentions, goals and values. Looking at social media posts, for example, we can understand consumers' opinions and emotions, or citizens' vote intentions. Crime investigations were solved by looking at words used by criminals, for example in ransom demands or other threat communications (Olsson & Luchjenbroers, 2014).

In the era of big data, an unimaginable amount of information has become available to scholars, including large text corpora and patterns of social interactions. In the meantime, the primary focus of social and behavioral research has shifted from the study of the attributes of social units to that of their relationships. "The fundamental difference between a social network explanation and a non-network explanation of a process is the inclusion of concepts and information on relationships among units in a study" (Wasserman & Faust, 1994). For this reason, social network analysis (SNA) has become a key discipline for the interpretation of many real world phenomena, in research and in business. The concept of "network" evolved and became flexible: scholars not only study relationships among people, groups and organizations, but also consider networks of words, concepts and ideas.

In such a scenario, the advantage of combining methods and tools of text mining and social network analysis is evident. Honest signals, which are subtle patterns of how we interact with other people (Pentland, 2008), are revealed by the study of language and of social networks, serving multiple goals. For example, actions which support successful relationships with clients, or improve employee communication, come from a better understanding of the impact that language use has within and across organizations (J Diesner, Frantz, & Carley, 2005; Fronzetti Colladon, Saint-Charles, & Mongeau, 2018; P. Gloor, Fronzetti Colladon, Giacomelli, Saran, & Grippa, 2017). The study of response times, rotating leadership and word use in email communication can reveal employees' disengagement and anticipate their intention to resign (Peter A. Gloor, Fronzetti Colladon, Grippa, & Giacomelli, 2017). Novel approaches in the analysis of networks of words support brand management in the era of big data (Fronzetti Colladon, 2018) and the forecasting of political elections (Fronzetti Colladon, 2019). Analogously, text mining helps brand managers to identify (virtual) consumer tribes (Peter A Gloor et al., 2019) and to develop customized marketing strategies (He, Zha, & Li, 2013). Business intelligence actions can also be supported by the combination of network and semantic analysis (e.g., Aswani, Kar, Ilavarasan, & Dwivedi, 2018).

While there are numerous studies on social network analysis and text mining in business, work at the intersection of these fields still lags behind in theoretical, empirical, and methodological foundations. In this special issue, we collect nine papers that all use text mining and network analysis, to present analytics, results and case studies, valuable to support decision-making processes in business.

Three papers leverage the information contained in social media, with the aim of profiling users, digging into their opinions and topics of interest. Wu, Li, Shen and He (2019) propose an opinion summarization method for microblog texts based on emotion modeling and deep learning. They use the Ortony-Clore-Collins model of emotion and convolutional neural networks, to automatically evaluate the sentiment of posts that appeared on Chinese microblogging systems. Their model is validated by comparing results with a manually annotated dataset. In the work of Greco and Polli (2019) customers are profiled through the analysis of their tweets, using the Emotional Text Mining approach. The procedure, applied to tweets about a well-known sportswear brand, is useful to identify user communities and categorize customers and potential customers in terms of product preferences, sentiment and representations of the brand. Similarly, Gloor, Fronzetti Colladon, de Oliveira and Rovelli (2019) develop a tool for the automatic categorization of virtual consumer tribes. They build a web app, named *Tribefinder*, able to reveal Twitter users' tribal affiliations, by analyzing the language of their tweets. In their case study, the authors consider three tribal macro-categories (alternative realities, lifestyle, and recreation), useful to investigate users' lifestyle, hobbies, and political or ideological orientations. The social behavior of tribe members is further analyzed through social network analysis.

Misuraca, Scepi and Spano (2019) also focus on customers and analyze tweets, but considering post-sale interactions. They present a methodology to automatically manage the information listed in the requests that customers send to the social media accounts of companies. Their procedure relies on the use of clustering and network techniques for extracting high-level structures from texts, considering relationships between concepts and co-occurring words. A similar goal is pursued by Celardo and Everett (2019), who review many approaches of text clustering and propose one based on the use of the Louvain algorithm and the study of one- and two-mode networks.

Lanningan (2019) looks at text from a different perspective and is interested in the comparison of online and face-to-face discourse dynamics. He uses networks and text mining to compare discourse employed in-person and on social media platforms of four specialty coffee events in Canada. He finds that language has an impact on the choice of coffee tastes and important implications for business. The author maintains that taste is not just an individual matter, even though it is in the individual that the combination of olfactory and discursive elements (used to explain sensory input) resides (Lannigan, 2019).

Chartier, Saint-Charles and Mongeau (2019) are interested in studying people preferences, expressed as journalists' inclination to use certain words in their writing. In their case study, the authors compare three different collaborative filtering algorithms applied to articles published by the New York Times. Results surprisingly show that, while similarities in actors' neighborhoods are a good predictor of semantic preferences, information on the individual social network adds little to the prediction accuracy.

The last two works in the issue (Aloini, Benevento, Stefanini, & Zerbino, 2019; Nolasco & Oliveira, 2019) extend the application fields of words and networks, showing that these techniques are extremely flexible. Aloini et al. (2019) use a combined approach, based on process mining, SNA and text mining, to improve process coordination dynamics in marine logistics. By adding text mining to traditional approaches they propose solutions for reducing process fragmentation and communication switching, which both negatively affect information exchange of port actors.

Lastly, Nolasco and Oliveira (2019) are interested in the study of the relationship between science and society and the impact they have on each other. They use topic modelling algorithms to extract and label social subjects and research themes, and then topic correlation metrics to create links among them. The efficacy of their approach is tested on a large Twitter corpus and on a PubMed article corpus, both about the Zika epidemic. Their results open new opportunities for the forecasting of social behaviors.

This is a special collection of heterogeneous and innovative research at the nexus of text mining and social network analysis. Our goal is to show the power of words and networks and solicit future research on this topic – while paying attention to the methodological issues arising when extracting networks from textual data (Jana Diesner,

2013), or when combining social network, discourse and semantic analysis (P A Gloor, 2017; Saint-Charles & Mongeau, 2018).

**Acknowledgements**

We wish to express our deep gratitude to all the authors and reviewers, for their contribution to this special issue, and to the Editor-in-Chief and the editorial office of the *International Journal of Information Management* for their invaluable support.